\documentclass{article}

\usepackage[preprint]{neurips_2025}

\usepackage[utf8]{inputenc} 
\usepackage[T1]{fontenc}    
\usepackage{hyperref}       
\usepackage{url}            
\usepackage{booktabs}       
\usepackage{amsfonts}       
\usepackage{nicefrac}       
\usepackage{microtype}      
\usepackage{xcolor}         
\usepackage{xspace}
\usepackage{wrapfig}
\usepackage[skins]{tcolorbox}
\usepackage{listings}
\usepackage{xcolor}
\usepackage{arydshln}
\usepackage{array}
\newcommand{\name}{GLARIFY\xspace}
\newcommand{\datasetone}{GLARIFY-Ambi\xspace}

\usepackage{subcaption}
\usepackage{graphicx}

\usepackage{amsmath,amsfonts,bm}

\def\eqref#1{equation~\ref{#1}}

\def\1{\bm{1}}

\DeclareMathAlphabet{\mathsfit}{\encodingdefault}{\sfdefault}{m}{sl}
\SetMathAlphabet{\mathsfit}{bold}{\encodingdefault}{\sfdefault}{bx}{n}

    \newcolumntype{Y}{>{\centering\arraybackslash}X}
    \usepackage{hyperref}
    \usepackage{url}
    \tcbset{
      aibox/.style={
        width=474.18663pt,
        top=10pt,
        colback=white,
        colframe=black,
        colbacktitle=black,
        enhanced,
        center,
        attach boxed title to top left={yshift=-0.1in,xshift=0.15in},
        boxed title style={boxrule=0pt,colframe=white,},
      }
    }
    \newtcolorbox{AIbox}[2][]{aibox,title=#2,#1}

\title{Resolving Ambiguity in Gaze-Facilitated Visual Assistant Interaction Paradigm}

\author{%
Zeyu Wang$^{1*}$, \quad
Baiyu Chen$^{2}\thanks{equal contribution}$, \quad
Kun Yan$^{3}$, \quad
Hongjing Piao$^{1}$, \quad \\
Hao Xue$^{2}$, \quad
Flora D. Salim$^{2}$, \quad
Yuanchun Shi$^{1}$, \quad
Yuntao Wang$^{1}\thanks{corresponding author}$ \\
\vspace{0.1cm}
$^{1}$ Key Laboratory of Pervasive Computing, Tsinghua University\\
$^{2}$ The University of New South Wales\\
$^{3}$ SKLSDE Lab,  Beihang University\\
\texttt{wang-zy23@mails.tsinghua.edu.cn}, \quad
\texttt{yuntaowang@tsinghua.edu.cn}
\vspace{1.0cm}
}

\begin{document}

\maketitle

\begin{abstract}

With the rise in popularity of smart glasses, users' attention has been integrated into Vision-Language Models (VLMs) to streamline multi-modal querying in daily scenarios. However, leveraging gaze data to model users' attention may introduce ambiguity challenges: (1) users' verbal questions become ambiguous by using pronouns or skipping context, (2) humans' gaze patterns can be noisy and exhibit complex spatiotemporal relationships with their spoken questions. Previous works only consider single image as visual modality input, failing to capture the dynamic nature of the user's attention. In this work, we introduce \name, a novel method to leverage spatiotemporal gaze information to enhance the model's effectiveness in real-world applications. Initially, we analyzed hundreds of querying samples with the gaze modality to demonstrate the noisy nature of users' gaze patterns. We then utilized GPT-4o to design an automatic data synthesis pipeline to generate the \datasetone dataset, which includes a dedicated chain-of-thought (CoT) process to handle noisy gaze patterns. Finally, we designed a heatmap module to incorporate gaze information into cutting-edge VLMs while preserving their pretrained knowledge. We evaluated \name using a hold-out test set. Experiments demonstrate that \name significantly outperforms baselines. By robustly aligning VLMs with human attention, \name paves the way for a usable and intuitive interaction paradigm with a visual assistant.

\end{abstract}

\section{Introduction} \label{sec:intro}

Recent advancements in Vision-Language Models (VLMs) have significantly enhanced AI systems' ability to interpret complex visual scenes and respond to natural-language queries about them. Concurrently, modern AR/VR headsets and smart glasses have integrated gaze-tracking capabilities~\cite{lv2024aria, tonsen2020high}, providing a natural attention signal that can be leveraged by such systems. This convergence has led to a growing interest in Human-Computer Interaction (HCI) research, which explores the interaction paradigm of gaze-facilitated visual question answering in everyday scenarios using eye-tracking glasses~\cite{wang2024g, konrad2024gazegpt, lee2024gazepointar}. These developments underscore the importance of incorporating gaze data into VLMs to enhance their adaptability and effectiveness in real-world applications.

VOILA-A~\cite{yan2024voila} addresses this challenge by utilizing a perceiver resampler module to integrate gaze trace heatmap features with single-image embeddings. However, this approach is limited to static scenarios with fixed viewing angles, making it inadequate for capturing the dynamic nature of user perception and attention. At the same time, recent advancements in VLMs have enabled the processing of video keyframes as input~\cite{xu2025qwen2, li2024llava, chen2024expanding}. To move forward, it is essential to account for the dynamic characteristics of users' gaze attention and their visual field, which present new challenges that cannot be resolved by simply mapping all gaze traces onto a single keyframe.

\begin{figure}
  \centering
    {\includegraphics[width=1\linewidth]{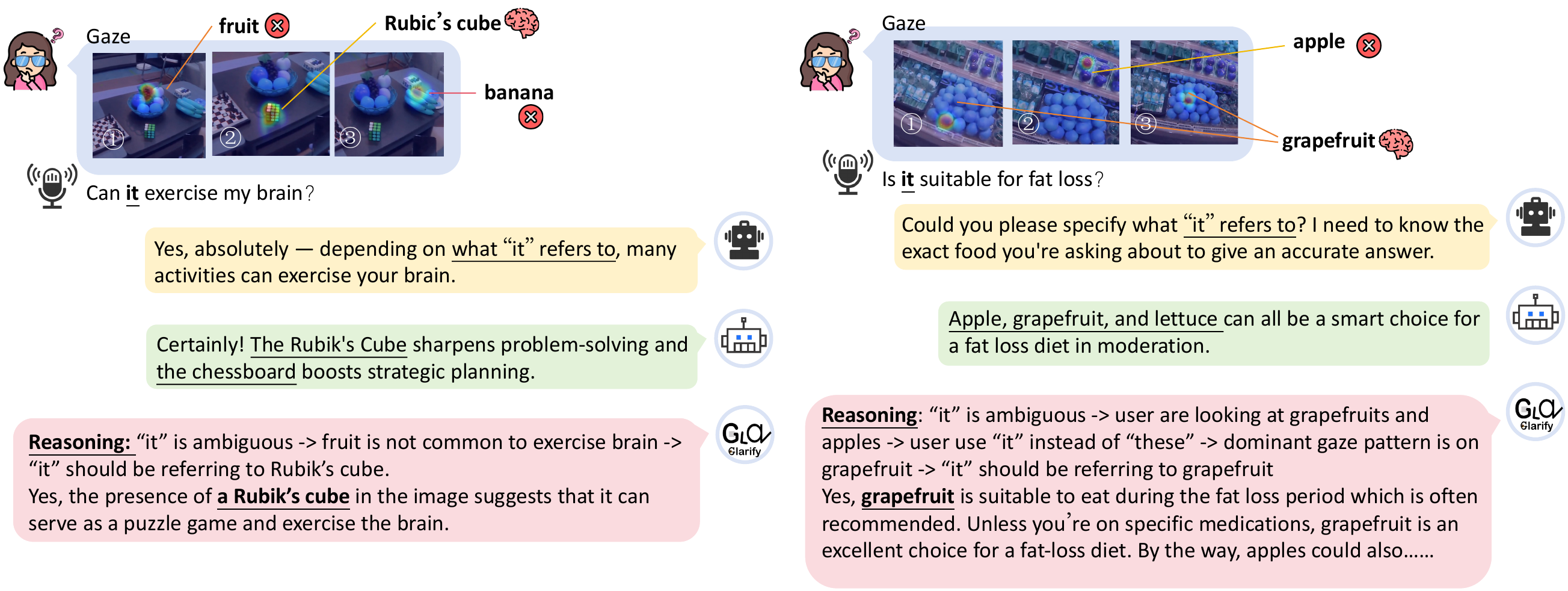}}
  \caption{Examples for gaze-facilitated visual querying in daily scenarios. We show the two ambiguous challenges in both examples.}
  \label{fig:teaser_examples}
\end{figure}

Specifically, two sources of ambiguity arise in real-world gaze-augmented interaction~\cite{wang2024g}. First, users' spoken queries often rely on pronouns or ellipses that lack explicit context description e.g., ``What is that made of?'', requiring additional grounding signals to resolve. Second, human gaze data is inherently noisy and temporally complex: users may glance at irrelevant objects, explore the scene non-linearly, or exhibit delays between visual attention and verbalization. As shown in Figure~\ref{fig:teaser_examples}, these factors hinder the direct use of gaze as a deterministic attention and instead necessitate robust modeling of its spatiotemporal patterns. We further illustrate it with a systematic analysis of gaze behavior collected from users wearing eye-tracking glasses in natural querying settings.

To address these challenges, we introduce \name, a novel method that robustly integrates dynamic gaze information into VLMs to support disambiguation in multimodal interaction scenarios. We first construct dataset \datasetone using GPT-4o~\cite{hurst2024gpt} within a carefully designed automatic data pipeline. Our pipeline simulates realistic, ambiguous language queries aligned with noisy gaze trajectories, which are approximated with a video mouse traces dataset~\cite{voigtlaender2023connecting}. This approximation has been justified by multiple works for saliency modeling~\cite{kim2017bubbleview, jiang2015salicon} and specifically for visual querying~\cite{yan2024voila}. We further design a mechanism to incorporate gaze's heatmap into a hierarchical VLM architecture. We conduct evaluations with a hold-out test set of \datasetone to validate our approach's effectiveness. 

Our contributions are threefold:
\begin{enumerate}
    \item We propose \name, a novel framework that resolves ambiguity in gaze-facilitated visual assistant interaction by robustly integrating spatiotemporal gaze signals into VLMs. We design a gaze-aware heatmap integration module that preserves pretrained VLM knowledge while effectively incorporating dynamic gaze features.
    \item We construct \datasetone, a synthetic dataset based on the VideoLN annotation with carefully aligned traces, ambiguous queries, and inserted noise, enabling supervised training of models to disambiguate intent from noisy attention signals. 
    \item We demonstrate our approach's effectiveness through evaluations on a hold-out test set. 
\end{enumerate}

\section{Related Works}

\subsection{Gaze-Enhanced Egocentric Vision Tasks}

Several recent works have begun incorporating eye gaze into vision-language models to better align them with human intent. For example, Voila-A~\cite{yan2024voila} augments a pre-trained VLM by incorporating user gaze heatmaps as an additional visual input, processed through ``Voila Perceiver'' modules to align the model’s internal attention with human fixations while preserving its original pretrained knowledge. G-VOILA~\cite{wang2024g} introduces a gaze-facilitated querying paradigm for AR settings, fusing users’ gaze, visual field, and voice to better understand their in-situ information needs. Its framework effectively integrates gaze data with the contextual information, yielding higher query accuracy compared to a baseline without gaze. In the domain of egocentric video QA for collaborative tasks, GazeVQA~\cite{ilaslan2023gazevqa} provides a dataset with recorded gaze and proposes AssistGaze, a model that grounds answers in gaze cues. Their work demonstrates that embedding gaze into the QA model is beneficial for resolving user intent in complex, task-oriented scenes, though its focus on structured industrial applications like assembly and disassembly may limit the direct applicability of its findings to more unconstrained daily activities. These studies collectively demonstrate the significant promise of using gaze to guide egocentric vision and QA. However, they generally assume that all recorded fixations are directly task-relevant and informative. A common limitation is the lack of explicit mechanisms to filter or reason about spurious gaze segments, such as those arising from user distraction, eye-tracker noise, or exploratory saccades unrelated to the immediate query, thereby leaving open the critical challenge of robustly handling noisy and temporally erratic gaze cues in continuous, real-world video streams, a gap our work aims to address.

\subsection{Attention Modeling in Vision–Language Models}

State-of-the-art vision–language transformers (e.g., LXMERT~\cite{tan2019lxmert}, ViLBERT~\cite{lu2019vilbert}, CLIP-ViL~\cite{shen2022how}) use multi-head attention for image-text alignment but typically rely on purely data-driven attention without natively incorporating human gaze or saliency. Early work~\cite{das2017human} introduced the VQA-HAT dataset and demonstrated that attention-based VQA models do not align well with human attention. Their analysis showed a significant discrepancy between machine-generated attention maps and the regions humans look at to answer visual questions. VQA-MHUG~\cite{sood-etal-2021-vqa} introduced a dataset with real eye-tracking data on both images and questions in VQA tasks. While it offered valuable comparisons between human and neural attention, it focused on static settings and diagnostic analysis, without incorporating gaze to improve model performance. To bridge this gap, some image-based methods inject human attention as supervision. For example, HINT~\cite{selvaraju2019taking} aligns a VQA model's gradient-based importance maps with human fixation maps, encouraging the network to ``look at'' the same regions as people. Similarly, Voila-A~\cite{yan2024voila} steers the VLM’s own attention to match gaze patterns. These techniques improve visual grounding and interpretability, but they have focused on static images or isolated queries. They do not perform temporal reasoning over attention sequences in videos. While some recent VideoQA models aim to better capture video dynamics by inferring importance from the video content and question itself, such as QueST~\cite{jiang2020divide} employing question-guided spatio-temporal attention, and TranSTR~\cite{li2023discovering} attempting to discover spatio-temporal rationales, they typically do not integrate explicit user gaze signals. Gaze could more directly indicate points of interest and help filter irrelevant information. In particular, existing models lack a mechanism to explicitly de-emphasize irrelevant or off-task fixations that occur over time.

\subsection{Video Question Answering without Gaze}

Most video QA systems ignore gaze entirely, relying only on visual and textual cues. Early datasets like TVQA~\cite{lei-etal-2018-tvqa} consist of hundreds of thousands of compositional QA pairs from TV show clips. It focused on temporal localization and dialogue fusion without considering user attention. Egocentric QA benchmarks, such as EgoVQA~\cite{fan2019egovqa} and EgoTaskQA~\cite{jia2022egotaskqa}, highlighted the challenges in understanding first-person visual inputs, including reasoning about actions, intents, and goals, yet neither incorporated gaze to resolve inherent ambiguities. Advanced egocentric video-language pre-training frameworks like EgoVLPv2~\cite{pramanick2023egovlpv2} have pushed performance by incorporating cross-modal fusion directly into the model backbones, leading to strong representations for various downstream tasks. More recent large vision-language models extended to video~\cite{zhang-etal-2023-video, li2023videochat, maaz-etal-2024-video, li2024llava, chen2024expanding, xu2025qwen2}, which have shown impressive capabilities in understanding video content and engaging in dialogue. However, these systems primarily infer relevance from the video content and language queries themselves, without explicitly leveraging eye gaze to disambiguate user intent or handle the temporal noise inherent in gaze data. In general, existing video QA methods implicitly assume that the relevant content can be inferred from the video alone. They neither leverage eye gaze to disambiguate intent nor explicitly handle the temporal noise of gaze data (e.g., brief saccades or fixations on irrelevant objects). As a result, model attention may be misaligned with the user’s true point of interest.

\subsection{Summary and Our Contribution}

Prior work has demonstrated that gaze can guide vision models~\cite{yan2024voila,wang2024g}, and some methods align model attention with human saliency~\cite{selvaraju2019taking}. However, no previous video QA approach explicitly filters out noisy gaze segments or uses multi-step reasoning to match gaze to question semantics. Our method fills this gap by explicitly detecting and discarding irrelevant fixations, then reasoning over the remaining gaze cues in a stepwise (chain-of-thought) fashion to align them with the question. By inserting gaze through lightweight adapters rather than re-training the entire VLM, we preserve the original pretrained performance~\cite{yan2024voila} while improving grounding in user intent. This combination of noise-robust gaze filtering and explicit reasoning sets our approach apart from existing work.zh

\section{Exploring the Noisy Nature of Human Gaze Pattern.} \label{sec:pattern}

In this section, we analyze how free-form questions and natural eye movements introduce uncertainty for a gaze-enabled visual assistant. In particular, we examined data from a prior work's user study~\cite{wang2024g}, where 21 participants were recruited to pose questions in three common daily scenarios using Pupil Labs Invisible smart glasses. 

The original G-VOILA analysis reported that \textit{``the majority of user queries involve usage of pronouns (68\%, 48\%, 43\%) and omission of certain constraint (20\%, 40\%, 39\%)''} and that participants \textit{``leverage gaze data to pinpoint objects of interest (67.2\%) and visual data to supplement context information (22.2\%)''}. This suggests that verbal questions for visual assistants are frequently ambiguous, validating our expectation that language-vision parsing is insufficient without additional visual grounding from gaze.

\begin{wrapfigure}[18]{t}{0.35\columnwidth}
    \centering
    \includegraphics[width=0.35\columnwidth]{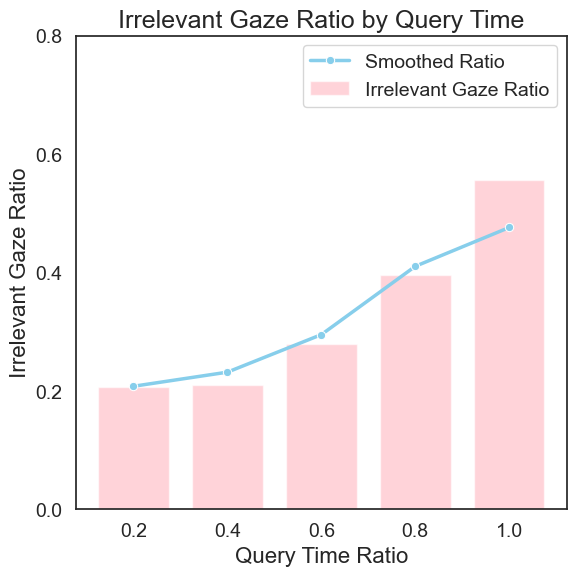}
    \caption{Gaze Ambiguity}
    \label{fig:gaze_ambiguity}
\end{wrapfigure}

To further investigate gaze noise, we analyzed each recorded query to compute the proportion of fixations that fell outside the user's intended referent region. We define an ``irrelevant gaze'' as a gaze point not spatially or semantically aligned with the user's querying intent. The irrelevant gaze ratio over time is given by:
\begin{equation}
R(t) = {\frac{N_{irr}(t)}{N_{total}(t)}}
\end{equation}
where $R(t)$ is the irrelevant gaze ratio at query time ratio $t \in [0, 1]$, $N_{irr}(t)$ and $N_{total}(t)$ is the number of irrelevant and total fixations in a temporal bin.
As shown in Figure~\ref{fig:gaze_ambiguity},the irrelevant gaze ratio remains consistently above 20\% throughout the querying process and rises sharply toward the end of queries, approximately 50\%. This pattern indicates that gaze data is inherently noisy and not uniformly aligned with user intent. For VLMs operating on key frames to perform video understanding, which is common in real-time, resource-constrained scenarios such as smart glasses, this poses a significant risk: key frames may coincide with off-target fixations. Therefore, models must robustly distinguish between relevant and irrelevant gaze traces to ensure accurate grounding and avoid misinterpretation of user intent.

\section{Methods}
\subsection{Automatic Data Synthesis Pipeline} \label{sec:method-datapipe}

\begin{figure}
  \centering
    {\includegraphics[width=1.0\linewidth]{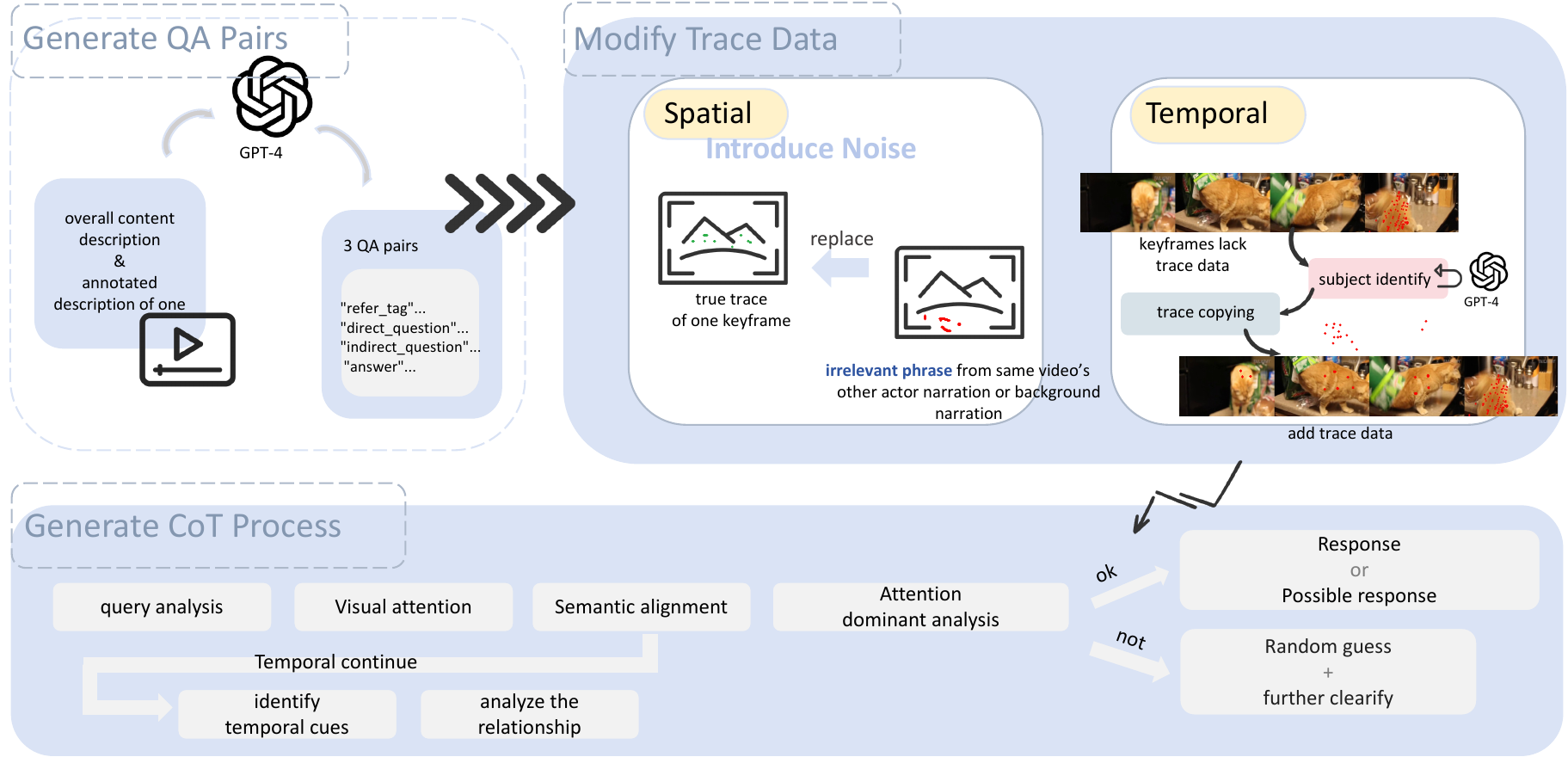}}
  \caption{Automatic Data Synthesis Pipeline.}
  \label{fig:pipeline}
\end{figure}

Inspired by VOILA-A~\cite{yan2024voila}, we leveraged mouse trace data for training, as gaze data is expensive to collect. The fairness and effectiveness of using mouse trace to mimic gaze trace have been justified by VOILA-A~\cite{yan2024voila} and other works~\cite{kim2017bubbleview, jiang2015salicon}.

\begin{table*}[!t]
    \centering
     \footnotesize
 \begin{tabular}{ccccccc}
\toprule
Split & Stage & \#Videos & \#Actors & \multicolumn{2}{c}{\#Questions} & SR \\ 
\cmidrule(lr){5-6}
& & & & Spatial & Temporal & \\
\midrule
Training & Generate QA pairs & 10287 & 25152 & \multicolumn{2}{c}{72912} & 92.79\% \\
Training & Modify Trace Data & 10272 & 25083 & 51134 & 21604 & 99.76\% \\
Training & Generate CoT & 10272 & 25081 & 51096 & 21164 & 99.34\% \\
Test & Generate QA pairs & 2126 & 5143 & \multicolumn{2}{c}{15013} & 93.80\% \\
Test & Modify Trace Data & 2123 & 5128 & 10429 & 4525 & 99.61\% \\
Test & Generate CoT & 2123 & 5127 & 10420 & 4440 & 99.37\% \\
\bottomrule
\end{tabular}
    \caption{Statistics of \datasetone Dataset, SR refers to QA pairs' Survival Rate from raw data after filtering}
    \label{tab:statistics}
\end{table*}

Our dataset builds on the Video Localized Narratives (VideoLN) dataset~\cite{voigtlaender2023connecting} Oops subset~\cite{epstein2019oops}, which provides multi-frame video annotations with simultaneous speech and mouse-trace grounding. Let $S = \{(V_i, N_i, T_i)\}$ be the video annotation from the VideoLN dataset, where $V_i = \{I_{i,1}, ..., I_{i,k}\}$ represents the keyframes for each video, $N^i=\{A_{i,1}, ..., A_{i,j}, A_{i,background}\}$ represent the actor-based narration, $T^i = \{T_{i,1}, ..., T_{i,j}, T_{i,background}\}$ is the trace corresponding to each narration and each points in the trace is assigned to one keyframe. For the three steps detailed below, we reserved a set of 200 samples for the pipeline design iteration. The prompt for the pipeline can be found in the appendix. The whole data pipeline was as shown in Figure~\ref{fig:pipeline}.

\subsubsection{Generate QA Pairs.} \label{sec:method_data_qa}
In this stage, we prompt GPT-4o to generate 3 QA pairs for each actor narration, except for background actor. Each QA pair should be about a segment of original narration and contains $(Q_{i,j}^D, Q_{i,j}^I, A_{i,j}, V_{i,j}, T_{i,j})$, where $Q_{i,j}^D$ and $Q_{i,j}^I$ represents the direct and indirect questions similar to~\cite{yan2024voila}, $A_{i,j}$ is the reference answer, $V_{i,j}$ is the narration's corresponding keyframes, $T_{i,j}$ is the narration segment's corresponding mouse trace. 

\subsubsection{Modify Trace Data.}
Though mouse trace could be used as gaze proxies, it does not have the noisy nature that human gaze does. To model realistic gaze uncertainty, we perturb the original mouse traces in two ways, corresponding to spatial and temporal reasoning cases.
\paragraph{Spatial correlation reasoning:} For QA pairs whose corresponding mouse trace $T_{i,j}$ has points for every keyframe, we introduce spatial ``noise'' by corrupting one frame’s trace. Concretely, we select one keyframe and replace its true trace with a misleading one: select an semantically irrelevant phrase from same video's other actor narration or background narration, then leverage the mouse trace corresponding to the phrase as the noise. This introduces the one advantage of using mouse trace data over gaze data: we can obtain the ground truth for irrelevant traces, thus promising a more reliable CoT process. 
\paragraph{Temporal correlation reasoning:} For QA pairs where some keyframes lack trace data (common in dynamic narratives), we simulate continuity of attention over time. We identify the subject of the narrative (e.g., the named actor or object) and copy its trace into the missing frames. This enforces a temporally consistent gaze following the event’s subject. In fact, the majority of these QA pairs refers to a dynamic event in the video, which poses the challenging task of pinpointing the user's interest event occurrences, especially with ambiguous reference in gaze-facilitated verbal questions. 
\subsubsection{Generate CoT Process.}
For each QA pair we also produce a step-by-step chain-of-thought (CoT) rationale, describing how one would use the trace to arrive at the answer. We tailor the CoT to the question type:
\paragraph{Spatial reasoning CoT:} Begin by parsing the question for ambiguous referents (e.g. pronouns like “he” or vague phrases). Next, analyze the trace segments across frames to see where attention was focused. Then check the semantic relevancy of the user's question and their visual targets: for example, the pronoun ``he'' should be referring to a man instead of certain object. For cases that user's referring question still cannot be clarified until this step, check if any entity clearly dominates user's visual attention and select it as the answer's basis.
\paragraph{Temporal reasoning CoT:} First perform the similar process for Spatial reasoning CoT to resolve QA pair's ambiguity. Then, identify any temporal cues in the question that imply an event sequence. Finally, analyze the relationship between the queried event and user's visual attention.

By following this structured CoT approach, we make explicit the role of user attention in reasoning. Each CoT is recorded as part of the dataset entry, effectively training models to consider intermediate reasoning steps. Finally, each synthesized data sample includes: $(Q_{i,j}^D, Q_{i,j}^I, A_{i,j}, V'_{i,j}, T'_{i,j}, CoT_{i,j})$. The final data composition is as shown in Table~\ref{tab:statistics}. During the evaluation, we randomly sample 2000 qa pairs in the test split and use the indirect question as the default setting.




\subsection{Model design}

We follow Voila-A's design principle to avoid introducing a significant number of new parameters due to the limited size of the \datasetone dataset. Similarly, this paper aims to prove that our proposed CoT process could lessen the ambiguity issue introduced by incorporating gaze modality to video VLMs. Therefore, the modification of our model is simplified to highlight the benefits brought by the robust reasoning process. 

\begin{figure}
  \centering
    {\includegraphics[width=0.8\linewidth]{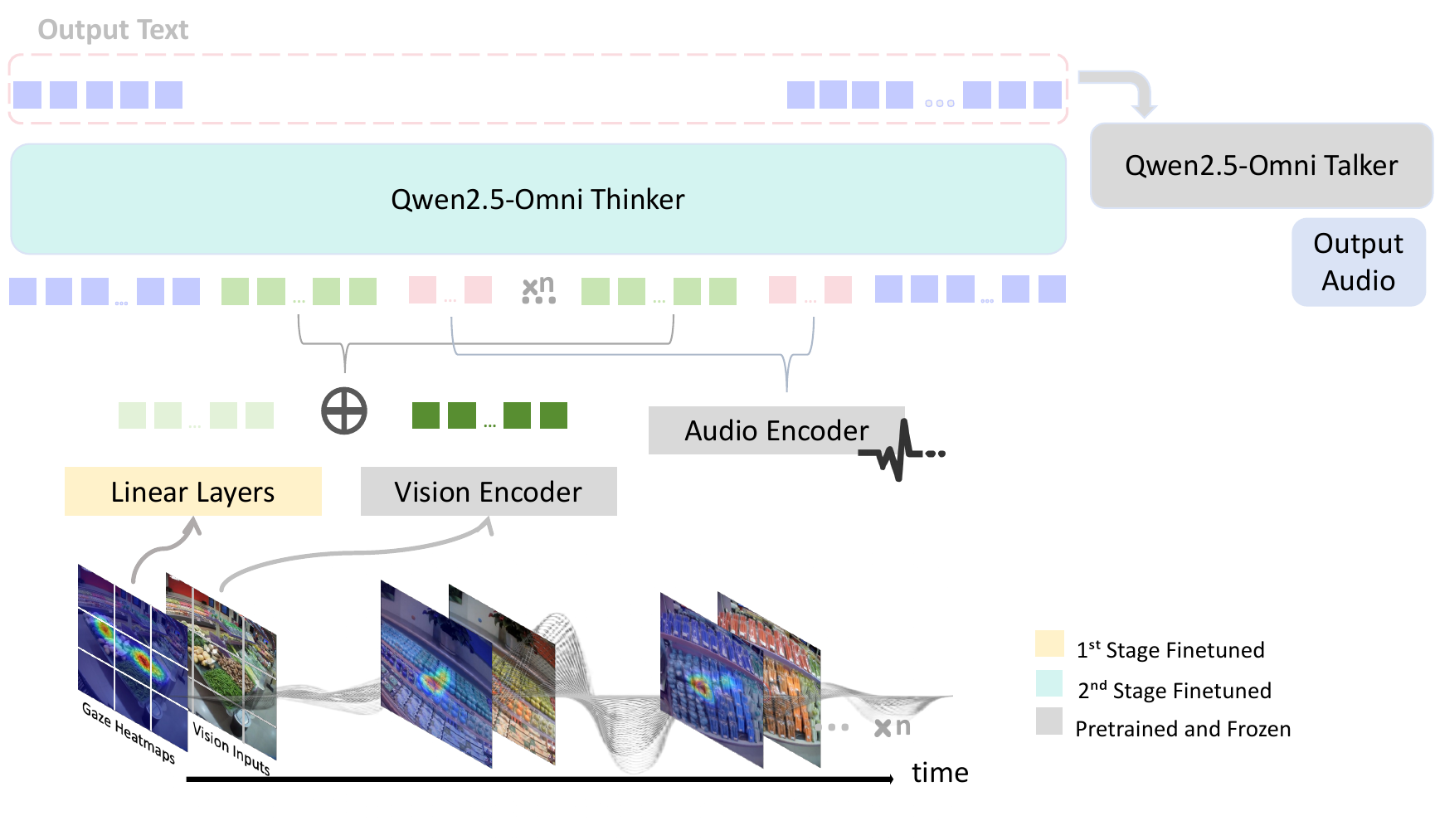}}
  \caption{Overall model architecture of \name. Gaze heatmaps and visual frames are processed jointly through linear projection and vision encoder, with fused tokens passed to Thinker for reasoning. The Talker generates textual and spoken outputs. Shading indicates modules trained at different stages: Stage 1 trains only the gaze projection layers, while Stage 2 fine-tunes the Thinker.}
  \label{fig:model}
\end{figure}

We adopt Qwen2.5-Omni-3B model~\cite{xu2025qwen2} as our based model architecture. As shown in Figure~\ref{fig:model}, Qwen2.5-Omni is a unified end-to-end multimodal model that take text, images, audio, and video as input and generates streaming text and speech outputs. These pretrained modules share a common encoder–decoder paradigm: the image/audio encoders map each input modality into a sequence of token embeddings, which are then fused in the Thinker via shared self-attention. Notably, our goal is to align the model's attention on images with the user's gaze attention, as well as the trace data has limited amount, thus training gaze as an encoder module that output independent token embeddings is unrealistic. Therefore, \name encodes gaze information and injects it into image embeddings.

Specifically, \name augments the vision branch with a heatmap module that encodes the user’s spatiotemporal gaze. Justification for using heatmap as gaze trace representation has been illustrated by~\cite{yan2024voila}'s dedicated ablation study. Let $I\in\mathbb{R}^{T\times C\times H\times W}$ denote the extracted video keyframes and $H\in\mathbb{R}^{T\times 1\times H\times W}$ the corresponding gaze heatmaps (one grayscale heatmap per keyframe, with values indicating user gaze attention density).  We process $H$ through the same patchify function as $I$.  Concretely, we define:

\begin{equation}
X = \mathrm{patchify}(I),\quad X\in\mathbb{R}^{T\times C\times H' \times W' \times p \times p},
\end{equation}
\begin{equation}
G = \mathrm{patchify}(H),\quad G\in\mathbb{R}^{T\times 1\times H' \times W' \times p \times p},
\end{equation}

where each keyframe is split into $H' \times W'$ patches and $p$ is the side length of each patch.  A small trainable linear layer $f_{\mathrm{linear}}:\mathbb{R}^{p\times p}\to\mathbb{R}^{D}$ is applied to the raw gaze tokens $G$ to produce gaze embeddings of the same size as image token embeddings:

\begin{equation}
Z = f_{\mathrm{linear}}(G),\quad Z\in\mathbb{R}^{T\times H' \times W'\times D}.
\end{equation}
\begin{equation}
V = ViT(X), \quad V\in\mathbb{R}^{T\times H' \times W'\times D}
\end{equation}

This projection aligns each gaze patch with its corresponding image patch's token embedding. The core integration in \name is an elementwise fusion of gaze and visual tokens.  For each keyframe and image patches, let image embedding $V_t\in\mathbb{R}^{D}$ be the original image patch tokens and $Z_t\in\mathbb{R}^{D}$ be the projected gaze tokens. We compute fused tokens by simple addition:

\begin{equation}
\widetilde{V_t} = V_t + Z_t, \quad t=1,...,T\times H'\times W'
\end{equation}

These fused vision tokens $\widetilde{V}$ are then concatenated with any other modality tokens (audio or text) and fed into the Thinker module of Qwen2.5-Omni for multimodal inference. This fusion biases the model’s attention toward gazed regions: adding $Z$ effectively raises the token values at patches the user fixates on. This approach retains the full visual context (all patches) rather than cropping or discarding peripheral regions~\cite{konrad2024gazegpt}. Furthermore, this incorporate strategy only introduces limited new parameters which is 0.0341\% of the total parameters in our implementation. This lightweight alignment method is more suitable for incorporating data modalities with only a limited amount of data and proven to be more data efficient compared to Q-formers or attention mechanisms~\cite {li2024llava, li2023blip}.

\subsection{Training}
We leveraged swift~\cite{zhao2024swift} training infrastructure to train \name. To preserve the pre-trained knowledge of Qwen2.5-Omni, we adopted a two stage training method, as shown in Figure~\ref{fig:model}. Only the heatmap projection layer $f_{\mathrm{linear}}$ are learned from gaze-labelled data generated from Step~\ref{sec:method_data_qa} without a reasoning process. This is can help the model to easily comprehend the spatial relationship between gaze heatmap and user's attention, compared to more complex reasoning text, which is further illustrated in Section~\ref{sec:exp_abla_train}. All original Qwen2.5-Omni modules (Vision Encoder, Audio Encoder, Thinker, Talker) are kept frozen during this stage. For the second stage, we train the thinker module and the projection layer $f_{\mathrm{linear}}$ jointly on \datasetone. The first stage is trained for 1 epoch, and the second stage is trained for 2 epochs.

\section{Experiments}

\subsection{Evaluation metrics}

\paragraph{Objective metrics -- GPT-Accuracy.} We leveraged Azure OpenAI Evaluation \footnote{https://ai.azure.com/resource/evaluation} to craft our automatic evaluation metrics. We evaluate the factual accuracy of a generated response by comparing to the answer generated through previous data synthetic pipeline. The response that is considered as factually aligned should be: (1) a consistent superset of the answer, (2) matched with all details of the answer, or (3) different from the answer yet the factual accuracy is unaffected. The reported number stands for the percentage of factually aligned response in the test set.

\paragraph{Objective metrics -- Video-Factuality.} GPT-Accuracy metric only considers the key points stated in the reference answers, which might expand to providing extra explainable information that is not necessary. Therefore, we randomly sampled 100 responses and required one annotator to score the responses based on the facts in the original video. 

\paragraph{Subjective metrics.} To holistically evaluate the quality of model-generated responses, we employ two user-centric subjective metrics: helpfulness and trust. These metrics are assessed through human evaluation, where annotators are instructed to adopt the perspective of a user seeking assistance.

\begin{itemize}
    \item \textbf{Helpfulness} measures the extent to which a response addresses the user’s query effectively. Annotators evaluate whether the answer resolves the stated problem, provides reasonable insights, and avoids redundancy or irrelevance. High helpfulness scores indicate that the response is comprehensive, contextually appropriate, and aligned with the user’s intent.
    \item \textbf{Trust} assesses the perceived reliability and credibility of the model’s output. Annotators consider factors such as logical coherence, factual accuracy, explainability, and consistency with domain-specific knowledge.
\end{itemize}

All scores provided by human evaluators are on a Likert scale (1-10).

\subsection{Main results}

\begin{table}[ht]
\centering

\begin{subtable}[c]{0.45\linewidth}
\centering
\setlength{\tabcolsep}{2pt}
\footnotesize 
\begin{tabular}{lc}
\toprule
Methods & GPT-Accuracy \\
\midrule
InternVL2.5-4B & 23.15\% \\
Qwen2.5-Omni-3B & 28.3\% \\
Qwen2.5-Omni-3B + Finetune & 26.7\%  \\
Qwen2.5-Omni-3B + Gaze & 29.85\% \\
\name (ours) & \textbf{38.05\%} \\
\bottomrule
\end{tabular}
\caption{GPT-Accuracy across different model variants.}
\label{tab:main_results}
\end{subtable}
\hfill
\begin{subtable}[c]{0.54\linewidth}
\setlength{\tabcolsep}{2pt}
\centering
\footnotesize
\begin{tabular}{lcccc}
\toprule
Methods & GPT-Acc. & Video-Fact. & Helpful & Trust \\
\midrule
Qwen2.5-Omni-3B & 28.3\% & 7.79 & 6.22 & 6.31 \\
\name stage1 & 28.5\% & 7.99 & 7.15 & 7.34 \\
\name cot only & 31.7\% & 7.84 & 7.19 & 7.44 \\
\name (ours) & \textbf{38.05\%} & \textbf{8.51} & \textbf{7.64} & \textbf{7.74} \\
\bottomrule
\end{tabular}
\caption{GPT-Accuracy and human evaluation of factuality, helpfulness, and trust for ablated variants of \name.}
\label{tab:result_pipeline}
\end{subtable}
\caption{
Main results of model performance.
}
\end{table}

The main results were as shown in Table~\ref{tab:main_results}. Our method, \name, achieves a GPT-Accuracy of 38.05\%, surpassing all baselines by a significant margin. The results validate our core hypothesis: modeling spatiotemporal gaze patterns and reasoning user attention jointly with query is critical for resolving ambiguities in gaze-facilitated visual queries.

Notably, the base Qwen2.5-Omni-3B model achieves 28.3\% accuracy, but its performance drops to 26.7\% after standard fine-tuning, highlighting the difficulty of resolving verbal ambiguity without the guidance of human gaze signal. Simply appending gaze features to the base model without reasoning process yields a modest improvement (29.85\%), underscoring the limitations of naive integration of noisy gaze signal. In contrast, the performance gap between \name and ``Qwen2.5-Omni-3B + Gaze'' confirms that our synthesized training data and chain-of-thought reasoning are essential for robust gaze integration.

\subsection{Training pipeline justification} 
\label{sec:exp_abla_train}

In this ablation study, we justified why the usage of QA pairs without a reasoning process is essential for \name. The method configuration ``\name Stage1'' refers to the model trained only with Stage 1 (heatmap projection layer on gaze QA pairs). ``\name CoT Only'' is a variant trained exclusively on \datasetone with chain-of-thought reasoning in both training stage. As shown in Table~\ref{tab:result_pipeline}, \name outperforms the variant trained solely on CoT data. This suggests that when introducing a novel modality like gaze heatmaps with limited training data, it is crucial to first establish elementary correspondences between the modality and textual outputs. The QA pairs provide simplified supervision that directly aligns spatial attention patterns with textual descriptions, whereas CoT reasoning requires simultaneous understanding of both modality-text relationships and multi-step logical inference.

\subsection{Qualitative results}
\begin{figure}
  \centering
    {\includegraphics[width=1.0\linewidth]{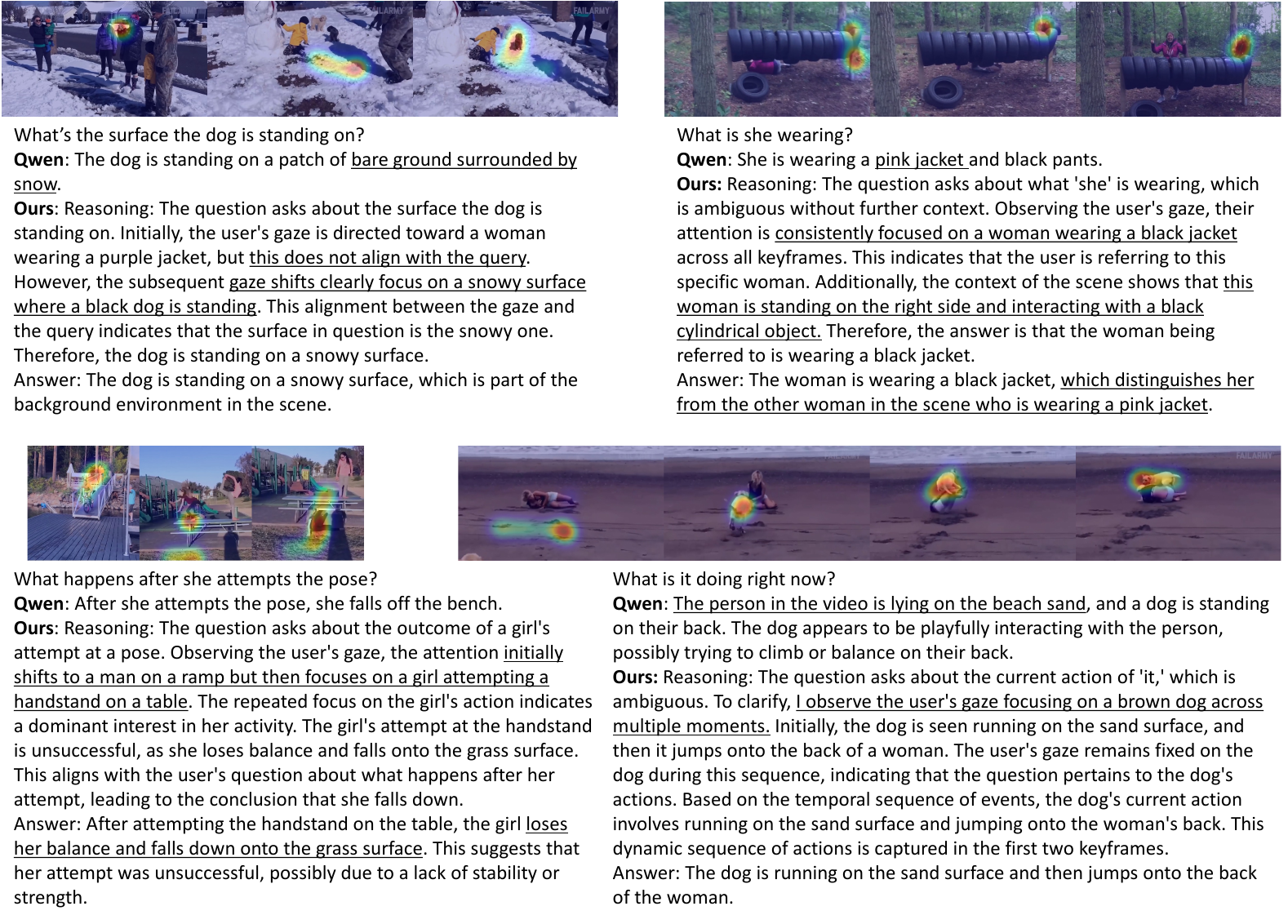}}
  \caption{Qualitative comparisons between \name and Qwen2.5-Omni-3B.}
  \label{fig:qualitative}
\end{figure}

To evaluate the effectiveness of our model, we conducted a qualitative analysis comparing its performance against the baseline model, Figure~\ref{fig:qualitative} shows some representative cases where \name effectively resolves ambiguous queries more accurately by aligning spatiotemporal gaze traces with user intent and applying structured reasoning to suppress irrelevant fixations.

\section{Conclusion}
In this work, we presented \name, a novel framework that addresses the critical challenge of ambiguity in gaze-augmented visual querying by integrating spatiotemporal gaze patterns into VLMs.  Our analysis revealed the inherent noise and complexity of real-world gaze signals, motivating the development of the \datasetone dataset, a scalable solution in resolving referential ambiguities and interpreting noisy attention signals. Experimental results demonstrate the effectiveness of \name and justify the training pipeline.

Future work should explore real-world deployment challenges, including temporal synchronization of gaze and visual data in streaming scenarios, and expansion to multilingual settings. By bridging the gap between human attention patterns and machine perception, \name represents a step toward seamless, gaze-enabled visual assistants in everyday environments.

\begin{ack}
Use unnumbered first level headings for the acknowledgments. All acknowledgments
go at the end of the paper before the list of references. Moreover, you are required to declare
funding (financial activities supporting the submitted work) and competing interests (related financial activities outside the submitted work).
More information about this disclosure can be found at: \url{https://neurips.cc/Conferences/2025/PaperInformation/FundingDisclosure}.

Do {\bf not} include this section in the anonymized submission, only in the final paper. You can use the \texttt{ack} environment provided in the style file to automatically hide this section in the anonymized submission.
\end{ack}

\bibliography{citation}

\begin{thebibliography}{10}

\bibitem{chen2024expanding}
Z.~Chen, W.~Wang, Y.~Cao, Y.~Liu, Z.~Gao, E.~Cui, J.~Zhu, S.~Ye, H.~Tian, Z.~Liu, et~al.
\newblock Expanding performance boundaries of open-source multimodal models with model, data, and test-time scaling.
\newblock {\em arXiv preprint arXiv:2412.05271}, 2024.

\bibitem{das2017human}
A.~Das, H.~Agrawal, L.~Zitnick, D.~Parikh, and D.~Batra.
\newblock Human attention in visual question answering: Do humans and deep networks look at the same regions?
\newblock {\em Computer Vision and Image Understanding}, 163:90--100, 2017.

\bibitem{epstein2019oops}
D.~Epstein, B.~Chen, and C.~Vondrick.
\newblock Oops! predicting unintentional action in video.
\newblock {\em arXiv preprint arXiv:1911.11206}, 2019.

\bibitem{fan2019egovqa}
C.~Fan.
\newblock Egovqa-an egocentric video question answering benchmark dataset.
\newblock In {\em Proceedings of the IEEE/CVF International Conference on Computer Vision Workshops}, pages 0--0, 2019.

\bibitem{hurst2024gpt}
A.~Hurst, A.~Lerer, A.~P. Goucher, A.~Perelman, A.~Ramesh, A.~Clark, A.~Ostrow, A.~Welihinda, A.~Hayes, A.~Radford, et~al.
\newblock Gpt-4o system card.
\newblock {\em arXiv preprint arXiv:2410.21276}, 2024.

\bibitem{ilaslan2023gazevqa}
M.~Ilaslan, C.~Song, J.~Chen, D.~Gao, W.~Lei, Q.~Xu, J.~Lim, and M.~Shou.
\newblock Gazevqa: A video question answering dataset for multiview eye-gaze task-oriented collaborations.
\newblock In {\em Proceedings of the 2023 Conference on Empirical Methods in Natural Language Processing}, pages 10462--10479, 2023.

\bibitem{jia2022egotaskqa}
B.~Jia, T.~Lei, S.-C. Zhu, and S.~Huang.
\newblock Egotaskqa: Understanding human tasks in egocentric videos.
\newblock {\em Advances in Neural Information Processing Systems}, 35:3343--3360, 2022.

\bibitem{jiang2020divide}
J.~Jiang, Z.~Chen, H.~Lin, X.~Zhao, and Y.~Gao.
\newblock Divide and conquer: Question-guided spatio-temporal contextual attention for video question answering.
\newblock In {\em Proceedings of the AAAI conference on artificial intelligence}, volume~34, pages 11101--11108, 2020.

\bibitem{jiang2015salicon}
M.~Jiang, S.~Huang, J.~Duan, and Q.~Zhao.
\newblock Salicon: Saliency in context.
\newblock In {\em Proceedings of the IEEE conference on computer vision and pattern recognition}, pages 1072--1080, 2015.

\bibitem{kim2017bubbleview}
N.~W. Kim, Z.~Bylinskii, M.~A. Borkin, K.~Z. Gajos, A.~Oliva, F.~Durand, and H.~Pfister.
\newblock Bubbleview: an interface for crowdsourcing image importance maps and tracking visual attention.
\newblock {\em ACM Transactions on Computer-Human Interaction (TOCHI)}, 24(5):1--40, 2017.

\bibitem{konrad2024gazegpt}
R.~Konrad, N.~Padmanaban, J.~G. Buckmaster, K.~C. Boyle, and G.~Wetzstein.
\newblock Gazegpt: Augmenting human capabilities using gaze-contingent contextual ai for smart eyewear.
\newblock {\em arXiv preprint arXiv:2401.17217}, 2024.

\bibitem{lee2024gazepointar}
J.~Lee, J.~Wang, E.~Brown, L.~Chu, S.~S. Rodriguez, and J.~E. Froehlich.
\newblock Gazepointar: A context-aware multimodal voice assistant for pronoun disambiguation in wearable augmented reality.
\newblock In {\em Proceedings of the 2024 CHI Conference on Human Factors in Computing Systems}, pages 1--20, 2024.

\bibitem{lei-etal-2018-tvqa}
J.~Lei, L.~Yu, M.~Bansal, and T.~Berg.
\newblock {TVQA}: Localized, compositional video question answering.
\newblock In E.~Riloff, D.~Chiang, J.~Hockenmaier, and J.~Tsujii, editors, {\em Proceedings of the 2018 Conference on Empirical Methods in Natural Language Processing}, pages 1369--1379, Brussels, Belgium, Oct.-Nov. 2018. Association for Computational Linguistics.

\bibitem{li2024llava}
B.~Li, Y.~Zhang, D.~Guo, R.~Zhang, F.~Li, H.~Zhang, K.~Zhang, P.~Zhang, Y.~Li, Z.~Liu, et~al.
\newblock Llava-onevision: Easy visual task transfer.
\newblock {\em arXiv preprint arXiv:2408.03326}, 2024.

\bibitem{li2023blip}
J.~Li, D.~Li, S.~Savarese, and S.~Hoi.
\newblock Blip-2: Bootstrapping language-image pre-training with frozen image encoders and large language models.
\newblock In {\em International conference on machine learning}, pages 19730--19742. PMLR, 2023.

\bibitem{li2023videochat}
K.~Li, Y.~He, Y.~Wang, Y.~Li, W.~Wang, P.~Luo, Y.~Wang, L.~Wang, and Y.~Qiao.
\newblock Videochat: Chat-centric video understanding.
\newblock {\em arXiv preprint arXiv:2305.06355}, 2023.

\bibitem{li2023discovering}
Y.~Li, J.~Xiao, C.~Feng, X.~Wang, and T.-S. Chua.
\newblock Discovering spatio-temporal rationales for video question answering.
\newblock In {\em Proceedings of the IEEE/CVF International Conference on Computer Vision}, pages 13869--13878, 2023.

\bibitem{lu2019vilbert}
J.~Lu, D.~Batra, D.~Parikh, and S.~Lee.
\newblock Vilbert: Pretraining task-agnostic visiolinguistic representations for vision-and-language tasks.
\newblock {\em Advances in neural information processing systems}, 32, 2019.

\bibitem{lv2024aria}
Z.~Lv, N.~Charron, P.~Moulon, A.~Gamino, C.~Peng, C.~Sweeney, E.~Miller, H.~Tang, J.~Meissner, J.~Dong, K.~Somasundaram, L.~Pesqueira, M.~Schwesinger, O.~Parkhi, Q.~Gu, R.~D. Nardi, S.~Cheng, S.~Saarinen, V.~Baiyya, Y.~Zou, R.~Newcombe, J.~J. Engel, X.~Pan, and C.~Ren.
\newblock Aria everyday activities dataset, 2024.

\bibitem{maaz-etal-2024-video}
M.~Maaz, H.~Rasheed, S.~Khan, and F.~Khan.
\newblock Video-{C}hat{GPT}: Towards detailed video understanding via large vision and language models.
\newblock In L.-W. Ku, A.~Martins, and V.~Srikumar, editors, {\em Proceedings of the 62nd Annual Meeting of the Association for Computational Linguistics (Volume 1: Long Papers)}, pages 12585--12602, Bangkok, Thailand, Aug. 2024. Association for Computational Linguistics.

\bibitem{pramanick2023egovlpv2}
S.~Pramanick, Y.~Song, S.~Nag, K.~Q. Lin, H.~Shah, M.~Z. Shou, R.~Chellappa, and P.~Zhang.
\newblock Egovlpv2: Egocentric video-language pre-training with fusion in the backbone.
\newblock In {\em Proceedings of the IEEE/CVF International Conference on Computer Vision}, pages 5285--5297, 2023.

\bibitem{selvaraju2019taking}
R.~R. Selvaraju, S.~Lee, Y.~Shen, H.~Jin, S.~Ghosh, L.~Heck, D.~Batra, and D.~Parikh.
\newblock Taking a hint: Leveraging explanations to make vision and language models more grounded.
\newblock In {\em Proceedings of the IEEE/CVF international conference on computer vision}, pages 2591--2600, 2019.

\bibitem{shen2022how}
S.~Shen, L.~H. Li, H.~Tan, M.~Bansal, A.~Rohrbach, K.-W. Chang, Z.~Yao, and K.~Keutzer.
\newblock How much can {CLIP} benefit vision-and-language tasks?
\newblock In {\em International Conference on Learning Representations}, 2022.

\bibitem{sood-etal-2021-vqa}
E.~Sood, F.~K{\"o}gel, F.~Strohm, P.~Dhar, and A.~Bulling.
\newblock {VQA}-{MHUG}: A gaze dataset to study multimodal neural attention in visual question answering.
\newblock In A.~Bisazza and O.~Abend, editors, {\em Proceedings of the 25th Conference on Computational Natural Language Learning}, pages 27--43, Online, Nov. 2021. Association for Computational Linguistics.

\bibitem{tan2019lxmert}
H.~Tan and M.~Bansal.
\newblock Lxmert: Learning cross-modality encoder representations from transformers.
\newblock In {\em Proceedings of the 2019 Conference on Empirical Methods in Natural Language Processing}, 2019.

\bibitem{tonsen2020high}
M.~Tonsen, C.~K. Baumann, and K.~Dierkes.
\newblock A high-level description and performance evaluation of pupil invisible.
\newblock {\em arXiv preprint arXiv:2009.00508}, 2020.

\bibitem{voigtlaender2023connecting}
P.~Voigtlaender, S.~Changpinyo, J.~Pont-Tuset, R.~Soricut, and V.~Ferrari.
\newblock Connecting vision and language with video localized narratives.
\newblock In {\em Proceedings of the IEEE/CVF conference on computer vision and pattern recognition}, pages 2461--2471, 2023.

\bibitem{wang2024g}
Z.~Wang, Y.~Shi, Y.~Wang, Y.~Yao, K.~Yan, Y.~Wang, L.~Ji, X.~Xu, and C.~Yu.
\newblock G-voila: Gaze-facilitated information querying in daily scenarios.
\newblock {\em Proceedings of the ACM on Interactive, Mobile, Wearable and Ubiquitous Technologies}, 8(2):1--33, 2024.

\bibitem{xu2025qwen2}
J.~Xu, Z.~Guo, J.~He, H.~Hu, T.~He, S.~Bai, K.~Chen, J.~Wang, Y.~Fan, K.~Dang, et~al.
\newblock Qwen2. 5-omni technical report.
\newblock {\em arXiv preprint arXiv:2503.20215}, 2025.

\bibitem{yan2024voila}
K.~Yan, Z.~Wang, L.~Ji, Y.~Wang, N.~Duan, and S.~Ma.
\newblock Voila-a: Aligning vision-language models with user's gaze attention.
\newblock {\em Advances in Neural Information Processing Systems}, 37:1890--1918, 2024.

\bibitem{zhang-etal-2023-video}
H.~Zhang, X.~Li, and L.~Bing.
\newblock Video-{LL}a{MA}: An instruction-tuned audio-visual language model for video understanding.
\newblock In Y.~Feng and E.~Lefever, editors, {\em Proceedings of the 2023 Conference on Empirical Methods in Natural Language Processing: System Demonstrations}, pages 543--553, Singapore, Dec. 2023. Association for Computational Linguistics.

\bibitem{zhao2024swift}
Y.~Zhao, J.~Huang, J.~Hu, X.~Wang, Y.~Mao, D.~Zhang, Z.~Jiang, Z.~Wu, B.~Ai, A.~Wang, W.~Zhou, and Y.~Chen.
\newblock Swift:a scalable lightweight infrastructure for fine-tuning, 2024.

\end{thebibliography}
\bibliographystyle{abbrv}


\appendix

\section{Hyperparameters}
Training was performed on 8 A800 GPUs using mixed precision (bfloat16) and Flash Attention for efficient memory usage. In Stage 1, the model was trained for 1 epoch with a batch size of 2 per device and gradient accumulation steps of 2, focusing on QA pairs without a reasoning process. In Stage 2, we resumed from the Stage 1 checkpoint, increased the batch size to 4 (with same accumulation steps), extended training to 2 epochs, and fine-tuned the entire thinker module on the final \datasetone dataset. A learning rate of 1e-4, gradient accumulation, and periodic evaluation and checkpointing were used throughout both stages.

\section{Prompts}
The prompts we used in our automatic data pipeline is as shown in Figure~\ref{fig:prompt1} and ~\ref{fig:prompt2}.

 \begin{figure}

    \begin{AIbox}{Prompt for Automatic QA Pairs Generation.}

    {\small \bf System Prompt:} Role: You are a visual assistant co-located with a user in an AR headset (HMD), jointly analyzing video keyframes. Your task is to generate QA pairs about the shared visual scene using both observed details and provided contextual descriptions.

\textbf{Input Content}

- Background information: provide overall content description of the video

- Referable sentence: annotated description of one actor/object/background in the video

\textbf{QA Generation Requirements}  

1. \textbf{Question Types}:  

   - detail question  
   
   - inference question that focus on Actions 
   
   - complex question requiring multi-step reasoning (Causal Chains/Temporal Analysis/Multi-Object Synthesis/Goal Inference/etc)

2. \textbf{Question Constraints}:  

   - Must have definitive answers within textual descriptions  
   
   - Each QA pair should have an indirect question that assumes a shared visual focus (e.g. using pronouns to refer to query target, or skipping context constraints)  

\textbf{Do Your Task Sequentially}  

Your task is to generate 3 QA pairs (include each question type once) based on the provided \textbf{Referable sentence}. Follow these steps:  

1. Copy \textbf{Referable sentence} and use <Q\#></Q\#> tags to extract all substrings that you want to query about, \# is the question number from 1–3.  

2. Generate a direct question and an indirect question for each extracted substring  

3. Provide a detailed answer to the direct question
   
\textbf{Format Rules}

- Avoid meta-references ("in the image/text") 

- Maintain AR perspective (use "ahead of us" instead of "in the frame") 

- JSON structure

\begin{lstlisting}
``` json
{
   "refer_content": str, original referable sentence with 3 <Q#></Q#> tags, for example: "<Q1>A man wearing shorts</Q1> is <Q2>performing gymnastics on a gymnastic bar</Q2> and then <Q3>gets unbalanced and falls down on the blue mat</Q3>.", tags can be nested if necessary,
   "qa_pairs": [
      {
         "refer_tag": str, tagged substring that you want to query about, for example "refer_tag" of the third question in the above example should be "<Q3>gets unbalanced and falls down on the blue mat</Q3>",
         "direct_question": "str, Full specification question about the refer_content",       
         "indirect_question": "str, Contextually reduced version, be as simple as possible, avoid using descriptive phrases, use pronouns instead and skip contextual description",         
         "answer": "str, response with details and reasoning, use multi-sentence response for complex question" 
      },
      ...
   ]
}
``` 
\end{lstlisting}
    \end{AIbox}
    \caption{Prompt for Automatic QA Pairs Generation.}
 \label{fig:prompt1}
    \end{figure}

 \begin{figure}

    \begin{AIbox}{Prompt for CoT Generation.}

    {\small \bf System Prompt:} Role: You are a visual assistant co‐located with a user in an AR headset (HMD), jointly analyzing video keyframes. Your task is to generate QA pairs about the shared visual scene using both observed details and provided contextual descriptions.

\textbf{Input Content}

- Background information: provide overall content description of the video

- Referable sentence: annotated description of one actor/object/background in the video

\textbf{QA Generation Requirements}  

1. \textbf{Question Types}:  

   - detail question  
   
   - inference question that focus on Actions  
   
   - complex question requiring multi‐step reasoning (Causal Chains/Temporal Analysis/Multi‐Object Synthesis/Goal Inference/etc)

2. \textbf{Question Constraints}:  

   - Must have definitive answers within textual descriptions  
   
   - Each QA pair should have an indirect question that assumes a shared visual focus (e.g.\ using pronouns to refer to query target, or skipping context constraints)  

\textbf{Do Your Task Sequentially} 

Your task is to generate 3 QA pairs (include each question type once) based on the provided \textbf{Referable sentence}. Follow these steps:  

1. Copy \textbf{Referable sentence} and use <Q\#></Q\#> tags to extract all substrings that you want to query about, \# is the question number from 1–3.  

2. Generate a direct question and an indirect question for each extracted substring  

3. Provide a detailed answer to the direct question
   
\textbf{Format Rules}  

- Avoid meta‐references (“in the image/text”)  

- Maintain AR perspective (use “ahead of us” instead of “in the frame”)  

- JSON structure

\begin{lstlisting}
```json
{
   "reasoning": str, you should stick to the reasoning core process, do not mention each step of the process, make the reasoning process flow naturally, 
}
``` 
 
\end{lstlisting}
    \end{AIbox}
    \caption{Prompt for CoT Generation.}
 \label{fig:prompt2}
    \end{figure}

\end{document}